\documentclass{article}
\usepackage[no-math]{fontspec}
\usepackage{iclr2027_conference,times}
\usepackage{amsmath,amssymb,booktabs,array,tabularx}
\usepackage{graphicx,xcolor,xurl}
\usepackage[hidelinks]{hyperref}
\newfontfamily\vizfont[BoldFont=texgyreheros-bold.otf]{texgyreheros-regular.otf}

\definecolor{vizink}{HTML}{293957}
\definecolor{vizblue}{HTML}{5271AE}
\definecolor{vizlight}{HTML}{70ACDE}
\definecolor{vizorange}{HTML}{FFA660}
\newcommand{\metric}[2]{#1\textsubscript{#2}}
\newcommand{\pname}[1]{P\textsubscript{#1}}
\newcommand{\tabstyle}{\vizfont\color{black}\fontsize{8}{10}\selectfont\setlength{\tabcolsep}{4pt}}
\title{The Default Trap: Rethinking Plan\\Evaluation in Tool-Using LLM Agents}
\author{%
Xueqi Li\textsuperscript{1}\thanks{Corresponding author: \texttt{xueqil@cs.cmu.edu}.}\qquad Jingjie Ning\textsuperscript{1}\qquad Yibo Kong\textsuperscript{1}\\
{\normalfont\textsuperscript{1}Carnegie Mellon University}\\
{\normalfont\texttt{\{xueqil,\,jening,\,yibok\}@cs.cmu.edu}}%
}
\iclrfinalcopy
\hypersetup{
  pdftitle={The Default Trap: Rethinking Plan Evaluation in Tool-Using LLM Agents},
  pdfauthor={Xueqi Li, Jingjie Ning, Yibo Kong}
}
\renewcommand{\iclrruler}[1]{}
\begin{document}
\raggedbottom
\maketitle
\fancyhead{}
\fancyhead[L]{Preprint}
\renewcommand{\headrulewidth}{0.4pt}

\begin{abstract}
An executor can respond strongly to a change in a supplied plan's priority while showing a small change in the same information-selection probability when a default-aligned whole plan is removed. We call the risk of interpreting the latter as weak responsiveness to alternative priorities the \emph{default trap}. We compare paired plans that prioritize different information targets with a shared no-plan reference. An accounting identity relates these distinct behavioral contrasts. Across 3,200 decision windows on 160 selected Retail, Airline, and AgentDojo tasks, switching priorities strongly redirects two models' choices, while the two plan-versus-default contrasts differ. In 2,160 additional windows, reversing account-list order shifts default target selection by 63.3--98.3 percentage points; priority-switching effects remain 96.7--100.0 points in either order. A separate 3,240-window component study finds strong control under single priority sentences, with effects of additional text varying by group and direction. Finally, 1,080 full-task episodes yield observed success differences of $-19.4$ to $+8.3$ points relative to no plan. All Retail and Airline success intervals include zero; AgentDojo results describe four fixed application worlds. These findings support joint reporting of priority responsiveness, presentation-dependent defaults, and task success and cost.
\end{abstract}

\section{Introduction}\label{sec:introduction}
Tool-using language agents use plans to translate goals into action priorities. A customer-support agent may inspect one order before another; a travel agent may compare hotel ratings before prices. Such guidance also connects planning and execution across multi-agent systems, whose performance depends on task structure \citep{cemri2025fail,kim2025scaling}. Understanding its role requires three linked measurements. \emph{Priority control}, also called directional control, is the change in action selection caused by changing a supplied priority. \emph{Default behavior} is execution without the supplied plan. \emph{Task utility} comprises task success and the resources used to achieve it.

An information target is a record or attribute to inspect through a tool. Consider an agent that reads A with an A-first plan and without a plan, but reads B with a matched B-first plan. The A-plan removal contrast is zero, while changing priority redirects selection. We call the risk of interpreting the first result as weak responsiveness to alternative priorities the \emph{default trap}. Whole-plan removal measures the change in a specified endpoint probability between plan-present and no-plan inputs; priority switching measures the response to replacing one priority with another. The first contrast alone does not determine the second. The no-plan reference can also depend on how information is presented.

\begin{figure}[t]
\centering
\includegraphics[width=\linewidth]{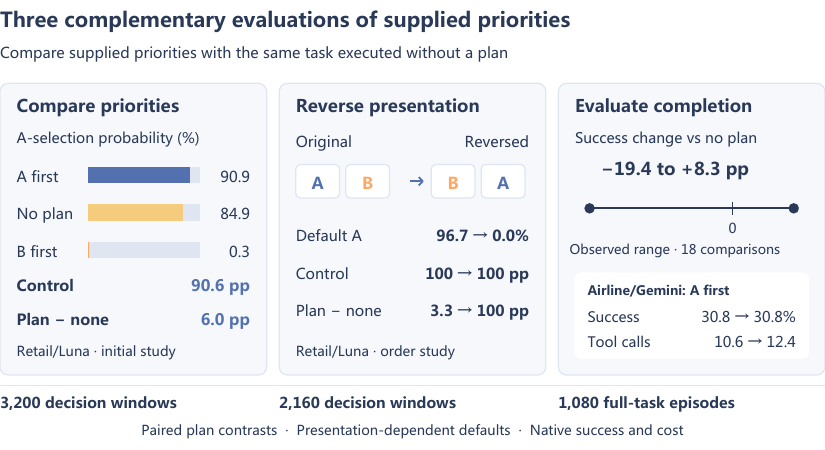}
\caption{\textbf{Three complementary evaluations of supplied priorities.} A and B are information targets. Control is A-selection probability under A-first minus B-first; the plan-minus-none contrast is A-first minus no plan. Effects marked pp are percentage points. Left and middle use Retail/Luna in separate samples; the middle arrow denotes account-list reversal with target identities fixed. Right shows 18 success point estimates across nine domain/model groups and an Airline/Gemini A-first example with sample success rates and mean attempted calls. The studies evaluate distinct interventions; the example's equal success point estimates do not establish equivalence. Decision windows stop at a measured read or stopping event; full-task episodes continue to termination or budget.}\label{fig:decomposition}
\end{figure}

We introduce a paired-priority framework at states with two useful \emph{reads}, tool calls that retrieve information in either order without changing the underlying business records. Each \emph{branch} identifies one target or query family. Paired plans retain shared guidance and replace a contiguous \emph{priority span}. A \emph{decision window} ends at the measured read or a recorded stopping event. The protocol compares both priorities with a shared no-plan reference and repeats identical inputs. An accounting identity connects the contrasts while retaining outcomes outside the two branches. In 3,200 windows from 160 selected customer-service and application-use tasks, two executor models show strong priority control alongside unequal plan-versus-default effects (Figure~\ref{fig:decomposition}).

Two further studies examine presentation and task outcomes. The \emph{list-order intervention} reverses account records while fixing target identities, plans, and business state. Across 2,160 windows with three models, default selection shifts by 63.3--98.3 points while control remains 96.7--100.0 points in both orders. The \emph{full-task evaluation} records local selection, benchmark-defined success, and resource use in the same 1,080 episodes. Strong local responsiveness accompanies uncertain completion gains. These studies use matched conditions and repeated executions, with separate task selections and stopping rules.

Our contributions are a matched, bidirectional measurement protocol with an accounting identity, a controlled test of how presentation changes the removal reference, and joint local and task-level measurements under supplied guidance. An additional component study compares full plans with single priority sentences and neutral guidance, delimiting what these measurements establish about plan structure.

\section{Related Work}\label{sec:related}
\textbf{Planning and execution.} ReAct interleaves reasoning and tool use \citep{yao2023react}; Plan-and-Solve decomposes reasoning tasks \citep{wang2023plansolve}. ReWOO separates reasoning from observations \citep{xu2023rewoo}, LLMCompiler schedules parallel calls \citep{kim2024llmcompiler}, and ADaPT decomposes failed subtasks \citep{prasad2024adapt}. Learned plans and reusable workflows also guide execution \citep{erdogan2025planandact,wang2025workflow}. Our protocol measures how such guidance changes action selection.

\textbf{Plan compliance and retention.} \citet{liu2026plantoaction} study plan removal, phase edits, reordering, reminders, compliance, and task success. \citet{mehta2026planspersist} examine retention through trajectory replay, internal representations, and context shortening. We study a narrower intervention: two permissible local priorities on matched inputs, a shared no-plan reference, and a presentation-by-priority cross. Plans remain visible throughout execution, so these comparisons measure input responsiveness rather than retention after eviction.

\textbf{Controlled attribution.} Reasoning perturbations measure output sensitivity \citep{lanham2023faithfulness}, and explanation studies identify influences omitted from stated reasoning \citep{turpin2023unfaithful}. \citet{ning2026revision} use four matched conditions to separate second-pass gains into re-solving, scaffold, and content contrasts. Their additive comparisons precede our accounting identity. Here the identity constrains the reporting of different plan-to-action contrasts; the empirical evidence comes from the implemented input interventions.

\textbf{Presentation and selection.} \emph{Position bias} means selection changes when an item's position changes with its content fixed. Studies document sensitivity to the order of prompt examples \citep{lu2022ordered}, effects of information placement in long inputs \citep{liu2024middle}, and tool preferences arising from names and descriptions, prior exposure, and list position \citep{blankenstein2026biasbusters}. Shopping agents show distinct positional responses in inspection and final choice \citep{wadi2026rank}; attention interventions examine tool-selection errors \citep{chen2026picking}. We cross account-record order with plan priority to measure how presentation shapes the default used in removal comparisons.

\textbf{Agent evaluation.} PlanBench evaluates planning \citep{valmeekam2023planbench}, WebArena tests interactive task completion \citep{zhou2024webarena}, and AgentBoard measures incremental progress \citep{ma2024agentboard}. Controlled planning comparisons also cover tasks coupling software decisions with physical processes \citep{decurto2026strategic}. We connect local control to success and cost using customer-service simulations from $\tau$-bench and $\tau^2$-Bench \citep{yao2025taubench,barres2025tau2bench} and AgentDojo application tasks \citep{debenedetti2024agentdojo}.

\section{Measuring Plan Influence}\label{sec:estimands}
\subsection{Matched inputs and exhaustive outcomes}\label{sec:definitions}
Each task fixes an initial state, conversation, tools, and two useful information targets A and B that can be read in either order. State preservation does not imply equal information value, cost, or downstream consequences. Plans $P_A$ and $P_B$ share their guidance and differ in one contiguous priority span. Each \emph{text condition}, a specific wording of the pair, defines three input conditions: $P_A$, $P_B$, and a shared no-plan condition $\varnothing$ that deletes the full plan field. We run each plan twice and no plan once, giving five separately initialized executions per wording. Each task has two wordings; their no-plan inputs are identical and provide two default draws. The repeat superscript in $P_k^{\rm repeat}$ denotes a fresh execution with identical inputs. For example, a frozen hotel plan prioritizes ratings or prices by changing only the corresponding word in its first sentence; Appendix~\ref{app:planexample} prints both spans and the complete shared text.

The endpoint $Y(z)$ records the decision-window outcome under input $z$. A and B denote reads of the corresponding target; OTHER denotes another read; BOTH denotes a read covering both targets. ERROR records a terminal tool or argument error, YIELD a response without a tool call, WRITE an action stopped before a state change or a call outside the approved read set, and BUDGET exhaustion of ten model responses. Every category remains in the analysis.

\subsection{Priority control and removal}\label{sec:decomposition}
Let $p_k(z)=\Pr[Y(z)=k]$ for branch $k\in\{A,B\}$, with task and text indices suppressed. The \emph{directional effects} measure the probability change from switching priority, and the \emph{removal contrasts} compare each matching plan with the default. We sign $R_k$ as plan minus no plan; the signed change on deleting the plan is $-R_k$,
\begin{equation}
\begin{aligned}
D_A&=p_A(P_A)-p_A(P_B), & D_B&=p_B(P_B)-p_B(P_A),\\
R_A&=p_A(P_A)-p_A(\varnothing), & R_B&=p_B(P_B)-p_B(\varnothing).
\end{aligned}\label{eq:contrasts}
\end{equation}
Adding and subtracting the default probability gives an accounting identity,
\begin{equation}
D_A=R_A+G_A,\qquad G_A=p_A(\varnothing)-p_A(P_B).\label{eq:decomposition}
\end{equation}
The \emph{default-to-B contrast} $G_A$ is the decrease in A selection when moving from no plan to the B plan. A small $R_A$ can accompany a large $D_A$ when default behavior already favors A. The default also sets positive \emph{headroom}, the largest possible increase in a branch's probability, through $R_k\leq1-p_k(\varnothing)$.

Let $p_O(z)=1-p_A(z)-p_B(z)$ collect the other six categories. Then
\begin{equation}
D_A=R_A+R_B+p_O(P_B)-p_O(\varnothing).\label{eq:nonbinary}
\end{equation}
This correction records changes in \emph{branch coverage}, the probability $C(z)=p_A(z)+p_B(z)$ of selecting exactly one designated target. The identity holds under common linear averaging over tasks and does not identify latent plan use or per-execution counterfactual transitions. Appendix~\ref{app:algebra} gives the symmetric relation and coverage checks.

\subsection{Repeatability and estimation}\label{sec:estimation}\label{sec:reference}
\emph{Repeat disagreement} is $N_k=\Pr[Y(P_k)\ne Y(P_k^{\rm repeat})]$, the probability that identical plan inputs yield different endpoint labels. It uses all eight categories. Each plan probability averages its two runs; the default uses its single run. We average text conditions within each task and give tasks equal weight. The follow-up studies average three repeats per task and condition. Effects are reported in percentage points and disagreement as a percentage.

For Retail and Airline, a \emph{customer-cluster bootstrap} resamples customers with all their tasks, conditions, and repeats together, preserving within-customer dependence. We report the middle 95\% of estimates from 10,000 resamples in the paired-priority study and 5,000 in each follow-up. AgentDojo estimates describe four fixed \emph{suite worlds}, saved application environments shared across tasks, covering travel, Slack messaging, workspace email, documents, and calendars, and banking. Appendix~\ref{app:algebra} records estimation details and the interpretation of empirical intervals.

\section{Experimental Design}\label{sec:design}
\subsection{Paired-priority study and matched plans}\label{sec:selection}\label{sec:plans}
We select states with two useful reads allowed by the benchmark rules whose order preserves business state. Retail contributes 96 tasks from 44 customers, Airline 28 tasks from 20 customers, and AgentDojo 36 benign tasks across four fixed application worlds. Here, \emph{benign} denotes ordinary task inputs without added adversarial instructions; \emph{native} refers to the benchmark's original data, rules, or scoring. A \emph{query configuration} groups tasks sharing a decision structure; the domains contain 51, 21, and 28 configurations. Retail comprises 20 archived dialogue cases, 40 cases from a previously held-out source group, and 36 cases screened for reorderable reads. The source label does not denote a fresh held-out evaluation here. Each source group receives fresh executions and a separate descriptive analysis.

Two text conditions per task provide 320 plan pairs. Plan preparation uses \mbox{\texttt{gpt-5.6-luna}} (Luna), with AI-assisted review. Each pair retains a shared body and edits one priority span. Five executions repeat each plan twice and execute no plan once, with identical initial state, history, tools, and continuation instructions. The archived Retail sample includes historical dialogues with 22 encrypted reasoning records retained as unreadable context; other Retail histories begin with a user message. Appendix~\ref{app:implementation} records preparation, source versions, and review provenance.

\subsection{Execution and follow-up controls}\label{sec:execution}
Plans enter through a \emph{developer message}, the application-supplied instruction channel, and remain in context. Removal deletes the complete plan. Retail and Airline receive plans before preparation; AgentDojo receives them after a verified \emph{prefix}, a saved sequence of reads and observations. The paired-priority executors are Luna and \mbox{\texttt{gpt-5.4-mini}} (Mini). They use the provider's medium reasoning-effort setting, a 4,096-token output cap, one tool call per response, and up to ten responses. Known writes and unreviewed actions stop before execution.

The component study, list-order intervention, and full-task evaluation add \mbox{\texttt{gemini-3.8-flash}}, abbreviated Gemini. All three use one full plan pair per task, an 8,192-token output cap, medium reasoning, and three repeats per condition. Mini prepares and reviews the pairs in separate calls. Selection covers decision structures and customer diversity before formal outcomes are observed, excluding archived Retail dialogues. Appendix~\ref{app:followups} records selection seeds, prior engineering tests, provider interfaces, and request settings fixed before execution.

\subsection{Account-list order intervention}\label{sec:orderdesign}
The list-order intervention selects account-list traversal states: 20 Retail tasks from 20 customers and 20 Airline tasks from 18 customers. A and B identify the first and last records in the original order or reservation list. A program completes permitted identity and account reads, after which we present the returned list in original or reversed order. The semantic IDs, plans, other fields, and business state stay fixed. Subsequent account reads preserve the assigned order. All 40 states enter the analysis regardless of the observed default response.

Crossing two list orders, three plan conditions, three executors, and three repeats yields $40\times2\times3\times3\times3=2,160$ windows. We measure default branch probabilities and $D_A,R_A,R_B$ within each order. The \emph{order interaction} is $D_A^{\rm original}-D_A^{\rm reversed}$, the change in directional effect across presentations. The intervention targets the displayed list and therefore supplies a direct test of the default's dependence on presentation.

\subsection{Full-task evaluation}\label{sec:completiondesign}
The full-task evaluation uses 15 Retail, 13 Airline, and 12 AgentDojo tasks. Eligibility requires benchmark-provided reference actions that can be replayed, deterministic state checks, and opening requests consistent with the scenario as reviewed by Mini. Five candidates are excluded before formal execution. Three plan conditions, three executors, and three repeats yield $40\times3\times3\times3=1,080$ episodes. Simulated state-changing actions are permitted, with budgets of 60 executor and 16 simulated-user responses. We record the first local endpoint and continue \mbox{through task termination.}

\emph{Task success} means satisfying the benchmark's native completion criteria. Retail and Airline check database state, required actions, and communicated information, using Mini for required \emph{natural-language assertions}, criteria expressed in text. AgentDojo checks task success from the \emph{trajectory}, the recorded actions and observations, or the final state. Abnormal endings such as refusals and budget exhaustion count as failures. Mini simulates the Retail and Airline user from the original scenario and visible dialogue, with plans and tool results hidden. Tool calls and separate executor and simulated-user token counts include failed episodes. Appendix~\ref{app:followups} details scoring and review.

\section{Priority Control and Default Behavior}\label{sec:results}
\subsection{Strong control with unequal removal effects}\label{sec:mainresults}
The paired-priority study compares whole-plan removal with responses to changed priorities. Table~\ref{tab:main} shows substantial directional effects in all six domain/model groups, alongside smaller A-directed removal effects. $D_A$ ranges from 67.36 to 99.11 points, while $R_A$ ranges from 5.73 to 24.11. The default-to-B contrast $G_A=D_A-R_A$ spans 47.92--89.29 points. Figure~\ref{fig:control} in Appendix~\ref{app:sensitivity} displays the customer-cluster intervals, including Retail's $D_A$ intervals of $[82.67,97.47]$ for Luna and $[65.75,86.61]$ for Mini.

\begin{table}[!htbp]
\centering
\caption{\textbf{Joint measurements of plan influence.} $n$ counts tasks. $D$ measures priority switching, $R$ plan-minus-no-plan contrasts, and $N$ repeated-run disagreement; subscripts identify the branch. Effects are percentage points and disagreement a percentage. All endpoint categories remain \mbox{in the denominator.}}\label{tab:main}
{\tabstyle
\begin{tabular}{@{}lrrrrrrr@{}}
\toprule
Domain / model & n & \metric{D}{A} & \metric{D}{B} & \metric{R}{A} & \metric{R}{B} & \metric{N}{A} & \metric{N}{B}\\
\midrule
Retail / Luna & 96 & 90.63 & 92.97 & 5.99 & 91.41 & 4.69 & 5.73\\
Retail / Mini & 96 & 76.82 & 79.17 & 5.73 & 76.30 & 13.02 & 13.54\\
Airline / Luna & 28 & 99.11 & 99.11 & 9.82 & 100.00 & 1.79 & 0.00\\
Airline / Mini & 28 & 92.86 & 91.96 & 24.11 & 91.96 & 8.93 & 14.29\\
AgentDojo / Luna & 36 & 67.36 & 59.72 & 18.75 & 48.61 & 4.17 & 5.56\\
AgentDojo / Mini & 36 & 68.75 & 65.97 & 20.83 & 57.64 & 5.56 & 18.06\\
\bottomrule
\end{tabular}}
\end{table}

Retail/Luna illustrates the default trap. A selection is 90.89\% under the A plan, 84.90\% without a plan, and 0.26\% under the B plan. Switching priority therefore redirects selection by 90.63 points, while removing the A plan changes it by 5.99. Equation~\ref{eq:decomposition} accounts for the difference as $90.63=5.99+84.64$ after rounding. The small A-plan contrast and the large priority-switching contrast answer different intervention questions; neither implies an unobserved large A-plan effect relative to no plan.

The same pattern appears in Airline/Luna. Its default A probability of 89.29\% leaves 10.71 points of positive removal headroom. The observed $R_A=9.82$ nearly fills this range, while $D_A=99.11$. Across the six groups, B-directed removal effects exceed A-directed effects. Reporting either removal direction alone therefore yields a different impression of the same executor's \mbox{responsiveness to priorities.}

\subsection{Coverage accounts for changes beyond target switching}
The full outcome vocabulary identifies another source of removal effects. In Airline/Mini, non-A/B probability falls from 28.57\% without a plan to 5.36\% under the B plan. Its $R_B=91.96$ combines a 68.75-point decrease in A selection with a 23.21-point gain in branch coverage. The second term is the net increase in designated-branch coverage, a marginal probability change. Table~\ref{tab:probabilities} and Figure~\ref{fig:outcomes} preserve \mbox{these outcome distributions.}

Coverage also explains differences between the two directional effects. AgentDojo/Luna has $D_A=67.36$ and $D_B=59.72$, a 7.64-point gap that equals the A-plan versus B-plan coverage difference. OTHER and BOTH can include useful reads, so coverage describes target specificity. Retaining these outcomes makes changes in selection and stopping behavior visible within the \mbox{same outcome population.}

\subsection{Repeatability and variation across tasks}\label{sec:repeatresults}\label{sec:heterogeneity}\label{sec:cases}
Directional control and repeatability give complementary information. Retail/Luna has $N_A,N_B$ of 4.69\% and 5.73\%, compared with 13.02\% and 13.54\% for Mini. AgentDojo's similar $D_A$ estimates for the two models accompany B-plan disagreement rates of 5.56\% and 18.06\%. Thus executors with similar aggregate control can differ in consistency under identical inputs, a distinction central to agent reliability \citep{rabanser2026reliability}.

Task composition changes effect size while preserving the default gap. The Retail sample expanded through the two-read screen has $D_A=75.69$ for Luna and 62.50 for Mini, compared with 98.75 and 81.25 in the archived dialogues. AgentDojo travel has 55.26 and 47.37 points, while Slack has 81.82 and 93.18. Giving equal weight to customers, groups of identical or similar inputs, or query configurations retains a minimum $D_A-R_A$ gap of 40.89 points. Appendix~\ref{app:sensitivity} reports the \mbox{source and weighting checks.}

Recorded trajectories clarify the behavior behind these averages. In an order-management case and a round-trip flight search, Luna follows either supplied priority, while default runs skip a profile query and choose A. A hotel-comparison case requests prices under all five arms. The first two cases show preparation and target choice responding together; the third shows a task-specific default persisting across priorities. Appendix~\ref{app:cases} provides the exact action sequences.

\section{Presentation and Task-Level Outcomes}\label{sec:taskresults}
\subsection{Presentation order changes the removal baseline}\label{sec:order}
\begin{figure}[!t]
\centering
\includegraphics[width=\linewidth]{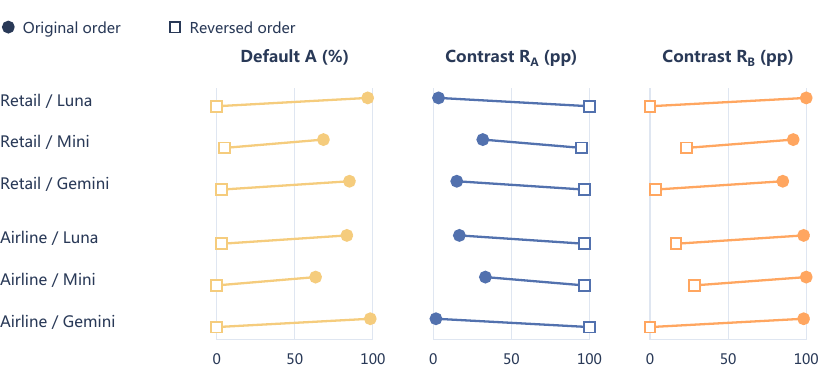}
\caption{\textbf{List-order reversal separates presentation-dependent defaults from supplied-priority control.} Each row contains 20 tasks and three repeats per condition. Left, default A selection under original and reversed account-list order. Center and right, A-directed and B-directed removal effects, with original order shown by filled circles and reversed order by open squares. Lines connect the two presentations. A-directed estimates are 100.0 points in both orders for five groups and 96.7 for Airline/Mini. Appendix~\ref{app:followups} reports paired intervals and the order interaction.}\label{fig:order}
\end{figure}

Figure~\ref{fig:order} summarizes all six domain and model groups in the list-order intervention. Reversing account-list order lowers default A selection by 63.3--98.3 points, with paired customer-cluster intervals strictly above zero for all six decreases. Retail/Luna moves from 96.7\% to 0.0\%, Retail/Mini from 68.3\% to 5.0\%, and Retail/Gemini from 85.0\% to 3.3\%. Airline shows the same direction, with reductions of 80.0, 63.3, and 98.3 points for Luna, Mini, and Gemini.

The A-directed effect is 100.0 points in both orders for five groups and 96.7 for Airline/Mini. Airline/Mini's order-interaction interval is $[-6.9,8.8]$ points; the other five intervals are $[0,0]$, as their observed effects are equal across all bootstrap resamples. In Retail/Luna, reversal moves $R_A$ from 3.3 to 100.0 points and $R_B$ from 100.0 to 0.0. Every group shows this reversal in the relative size of the two removal effects. Their changes follow algebraically from the associated branch probabilities. The experimental evidence is the response to the manipulated list order, which makes the default reference distribution part of the evaluation design.

This result extends the paired-priority study's distributional explanation with a controlled input change. Semantic targets retain their identities as their displayed positions move. An ablation report can therefore change substantially when the same records are presented in a different order, even as paired priorities yield strong control in each presentation. Reporting both directions makes this \mbox{dependence directly inspectable.}

\subsection{Local control and complete-task success}\label{sec:completion}
\begin{figure}[!htb]
\centering
\includegraphics[width=\linewidth]{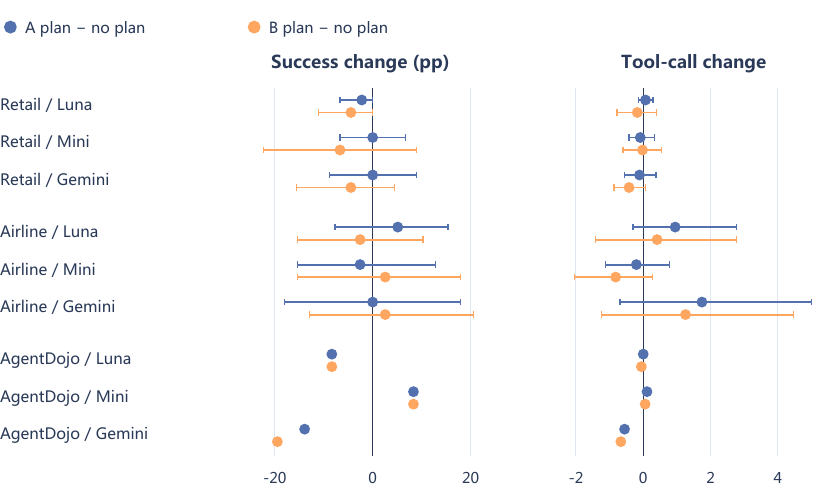}
\caption{\textbf{Task-success and tool-cost changes relative to no plan.} Blue and orange show A- and B-plan contrasts, in percentage points and mean calls per episode. Retail and Airline whiskers show 95\% intervals from 5,000 paired customer-cluster resamples. AgentDojo points describe \mbox{four fixed worlds.}}\label{fig:completion}
\end{figure}

The full-task evaluation measures local selection and eventual success on the same 1,080 episodes. Table~\ref{tab:completion} combines $D_A$ at the first endpoint with native success and mean attempted tool calls. Local control reaches 82.2--100.0 points in Retail and Airline. Their plan-versus-default success changes range from $-6.7$ to $+5.1$ points, and all 12 customer-cluster intervals include or touch zero (Figure~\ref{fig:completion}). These intervals leave practically important benefits and harms unresolved and do not establish equivalence.

\begin{table}[!htb]
\centering
\caption{\textbf{Local control and task outcomes in the same episode collection.} $D_A$ is the A-selection probability under A-first minus B-first, in percentage points. None means no plan; success is a percentage; cost is mean attempted tool calls, including failed episodes. Retail, Airline, and AgentDojo contain 15, 13, and 12 tasks, with three repeats per condition. AgentDojo describes \mbox{four fixed worlds.}}\label{tab:completion}
{\tabstyle
\begin{tabular}{@{}lrrrrrrr@{}}
\toprule
& & \multicolumn{3}{c}{Success (\%)} & \multicolumn{3}{c}{Tool calls}\\
\cmidrule(lr){3-5}\cmidrule(l){6-8}
Domain / model & \metric{D}{A} & None & A plan & B plan & None & A plan & B plan\\
\midrule
Retail / Luna & 93.3 & 80.0 & 77.8 & 75.6 & 6.6 & 6.7 & 6.4\\
Retail / Mini & 82.2 & 75.6 & 75.6 & 68.9 & 6.6 & 6.5 & 6.6\\
Retail / Gemini & 97.8 & 84.4 & 84.4 & 80.0 & 8.1 & 8.0 & 7.7\\
\addlinespace[3pt]
Airline / Luna & 94.9 & 61.5 & 66.7 & 59.0 & 8.4 & 9.3 & 8.8\\
Airline / Mini & 94.9 & 61.5 & 59.0 & 64.1 & 8.5 & 8.3 & 7.7\\
Airline / Gemini & 100.0 & 30.8 & 30.8 & 33.3 & 10.6 & 12.4 & 11.9\\
\addlinespace[3pt]
AgentDojo / Luna & 91.7 & 83.3 & 75.0 & 75.0 & 4.3 & 4.3 & 4.2\\
AgentDojo / Mini & 83.3 & 75.0 & 83.3 & 83.3 & 4.2 & 4.3 & 4.2\\
AgentDojo / Gemini & 33.3 & 100.0 & 86.1 & 80.6 & 6.3 & 5.8 & 5.6\\
\bottomrule
\end{tabular}}
\end{table}

The joint measurements make the distinction concrete. Retail/Luna has $D_A=93.3$ points, with success rates of 80.0\% without a plan, 77.8\% under A, and 75.6\% under B. Airline/Gemini has $D_A=100.0$ points, while success is 30.8\%, 30.8\%, and 33.3\%. Supplied priorities reliably change the initial information choice in these groups; task success additionally incorporates subsequent decisions, tool use, and termination.

AgentDojo provides a wider descriptive range on 12 tasks across four fixed worlds. Luna's $D_A=91.7$ accompanies success changing from 83.3\% to 75.0\% under either plan. Mini's $D_A=83.3$ accompanies an increase from 75.0\% to 83.3\%. Gemini has $D_A=33.3$, substantial OTHER probability, and success of 100.0\% without a plan, 86.1\% under A, and 80.6\% under B. Across all nine groups, the 18 success changes span $-19.4$ to $+8.3$ points. The direct success contrast $P_B-P_A$ ranges from $-7.7$ to $+5.1$ points in Retail and Airline, with all six customer intervals including zero (Appendix~\ref{app:prioritysuccess}). This compares the persistent priority texts; because plans remain visible, it does not isolate the effect of the first observed read.

\subsection{Resource use and completion timing}
Tool and token counts complement success. Airline/Gemini averages 10.6 calls without a plan, 12.4 with A, and 11.9 with B; both call-difference intervals include zero. Mean executor tokens rise from 90.7 thousand to 133.8 and 138.5 thousand, increases of 47.5\% and 52.6\%. The two absolute token-difference intervals are positive under the reported customer bootstrap (Appendix~\ref{app:cost}). These comparisons include failed episodes and follow the provider's token accounting. AgentDojo/Gemini uses fewer calls under both plans alongside lower observed success, illustrating why calls alone do not measure efficiency.

\emph{Cumulative success at a response threshold} is the fraction of episodes that have ended successfully by that many executor responses. In Airline/Gemini, the ordering of the three conditions changes between ten responses, twenty responses, and the full budget. These are completion-time summaries of the same trajectories, not effects of assigning a different execution budget; Appendix~\ref{app:cost} reports the values.

\subsection{Component controls}\label{sec:componentresults}
A separate component study adds 3,240 windows on 60 selected states (Appendix~\ref{app:components}). Single priority sentences produce $D_A=86.7$--$100.0$ points in Retail and Airline. Adding the surrounding plan text changes control differently across groups and directions: Retail/Mini's $D_B$ is 10.0 points lower with the full plan, while neutral guidance also changes endpoint probabilities relative to no plan. These input contrasts bound the interpretation of local control; they neither establish equivalence nor assign a unique contribution to multi-step planning, shared semantics, or text length.

\section{Implications for Plan Evaluation}\label{sec:discussion}
\textbf{Treat the default as an observed reference.} A plan aligned with a high-probability default has little positive removal headroom. The list-order intervention shows that this reference can move when only account-list presentation changes. Record semantic targets, presented order, and the default distribution alongside both directional removal comparisons.

\textbf{Match claims to the intervention.} A priority edit measures responsiveness to the supplied priority text. Whole-plan removal also changes preparation guidance, reminders, and text length. The identity relates these contrasts without making them interchangeable. The component study shows why a large priority response alone cannot establish the value of the surrounding plan structure.

\textbf{Evaluate control together with task outcomes.} Continued execution measures completion and resource use under each input condition. Reporting these outcomes alongside local selection distinguishes behavioral responsiveness from practical value. Extending the protocol to later decisions, richer plans, and multi-agent coordination is a direction for evaluation, beyond the systems tested here.

\section{Conclusion}\label{sec:conclusion}
We introduced a paired-priority framework that compares two supplied priorities with a shared no-plan reference and retains all measured outcomes. Its accounting identity distinguishes the questions answered by priority switching and whole-plan removal. The initial study finds strong directional control with unequal removal contrasts; the list-order intervention shows that the default reference depends on record presentation while priority control remains strong in the tested states.

The component study shows that single priority sentences can also elicit strong control, while additional guidance has group- and direction-dependent effects. Full-task episodes pair local measurements with completion and resource use. Observed success differences span $-19.4$ to $+8.3$ points, with customer-domain intervals including zero and AgentDojo limited to fixed-world descriptions. The supported conclusion is a reporting principle: evaluate responsiveness, defaults, and task outcomes together, while keeping their intervention targets and inferential limits explicit.
\label{main-text-end}

\clearpage
\subsection*{AI use statement}
Generative AI assisted plan preparation, manuscript drafting and cross-review, literature review, vector graphics, and offline verification. The paired-priority study used Luna-assisted plans and AI-assisted curation. The component study, list-order intervention, and full-task evaluation used Mini for plan preparation and separate review calls; full-task evaluation also used Mini for user simulation, required natural-language assertions, and review of compliance with benchmark rules, termed \emph{policy review}. The reviewer received trajectories with executor identity, plan text, and condition labels hidden. These are model-based judgments. Mini's multiple roles permit correlated errors, which the deterministic native checks and retained review records help examine.

\subsection*{Reproducibility statement}
The artifact contains source and task provenance, outcome indices, task-level results, complete condition and contrast data, vector-figure code, and numerical checks. The initial study and three follow-ups have separate source manifests and reuse selected tasks. The reported evidence comprises 8,600 decision windows and 1,080 full-task episodes; these execution counts are not counts of distinct tasks. Protocols, selection rules, model settings, native scoring, review uncertainty, and recovery are described in the appendix. Complete tables are supplied in \path{data/}; \path{data/README_ZH.md} indexes the condition, contrast, and component files. The complete paired-plan example appears in Appendix~\ref{app:planexample}. Frozen requests, native states, full trajectories, raw responses, and review artifacts are retained in the experiment directory.
\par
\begingroup
\raggedright
\def\UrlBreaks{\do\/}
\bibliography{references}
\bibliographystyle{iclr2027_conference}
\endgroup
\clearpage
\appendix

\section{Paired-Priority Intervention and Execution Details}\label{app:implementation}
\subsection{Source snapshots and selection}
Retail and Airline use the local $\tau^2$-Bench checkout \path{2174a603f6d014ef94473ffa95957f6ce27100db}. AgentDojo uses version 1.2.2 at checkout \path{089ed468cf3ed0322acc66b0211f26d9d90dbf60}. The adapters, interfaces connecting benchmark tools to the executor, preserve native business state and policy or system instructions. The selected endpoint is defined \mbox{before formal execution.}

Retail's structural expansion retains 36 of 39 reviewed candidates. Tasks 17, 60, and 61 lack the intended pair of requested targets; 15 earlier structural exclusions remain outside the expansion. AgentDojo retains 19 of 20 travel, 11 of 21 Slack, four of 40 workspace, and two of 16 banking tasks. Excluded cases include serial dependencies and cases in which one read already supplies the relevant information. The task manifests preserve all included IDs, branch definitions, provenance, \mbox{and grouping keys.}

\subsection{Plan preparation and the frozen input}
The final Retail and Airline revision corrects contradictory content in Retail task 112, residual order language in Airline task 33, and a shared preference in Retail task 7, each in its first text condition. AgentDojo's 72 pairs receive AI-assisted review before the formal batch. The review ledger separates this curation from model-audit status, with 53 revised and 19 retained pairs. The procedure combines structural span checks and AI-assisted inspection. Shared wording remains part of the implemented treatment, and the retained paired texts support \mbox{further independent assessment.}

The fixed input message contains the continuation instruction ``Continue helping this customer from the current conversation.'' Plan arms also contain the final plan text. The no-plan arm contains the continuation instruction alone. All 124 Retail and Airline cases have an empty saved prior-action list. AgentDojo replays its saved prefix before adding this message and verifies both returned observations and business state. The archived Retail encrypted history objects are preserved identically across arms and treated \mbox{as opaque context.}

\subsection{Executor settings and stopping rules}
Requests use the Responses API for model generation, with \path{reasoning.effort=medium}, \path{max_output_tokens=4096}, \path{store=false}, and \path{parallel_tool_calls=false}. Temperature, top-p, sampling seed, and tool choice are left at the service defaults. The tool inventories contain 16 Retail tools, 14 Airline tools, and 28/11/24/11 tools for travel/Slack/workspace/banking. Section~\ref{sec:design} identifies the executor models.

Each model response permits one tool call. Multiple calls terminate as ERROR before execution; a response with no call yields YIELD. Known writes and AgentDojo actions outside the reviewed whitelist of permitted read tools terminate as WRITE before execution. A target-family tool failure terminates as ERROR. A failure in preparation, such as an identity or account lookup, can return an observation and allow execution to continue. The loop ends at ten model responses if earlier stopping conditions have not been reached. Infrastructure failures leave an incomplete job for explicit recovery. In Slack, state checks exclude web-access logs from business state. The whitelist treats marking an email read as \mbox{a state change.}

Retail endpoints match order or product identifiers within the target tool family. Airline reservation endpoints match reservation IDs; flight endpoints match origin, destination, and date. Outbound queries enforce the requested nonstop constraint, and return queries permit a valid direct or one-stop search. AgentDojo uses the frozen query semantics. Attribute queries preserve the requested attribute, object queries match the intended object, document and calendar queries use returned identifiers, and webpage queries normalize URL schemes. A single read covering both targets receives BOTH. These rules assign \mbox{each endpoint label.}

\section{Paired-Priority Statistical Checks and Data Traceability}\label{app:sensitivity}
\subsection{Algebra and estimation details}\label{app:algebra}
For either branch, probability bounds give $-p_k(\varnothing)\leq R_k\leq1-p_k(\varnothing)$, which specifies the full removal range. The symmetric decomposition is $D_B=R_A+R_B+p_O(P_A)-p_O(\varnothing)$. Subtracting it from Equation~\ref{eq:nonbinary} yields
\begin{equation}
D_A-D_B=p_O(P_B)-p_O(P_A)=C(P_A)-C(P_B).\label{eq:coverage}
\end{equation}
Thus unequal directional effects exactly track unequal branch coverage. With exclusively A/B outcomes, both equal $R_A+R_B$. For Retail/Luna, the measured nonbinary decomposition is $90.63=5.99+91.41-6.77$ points after rounding. The negative correction records fewer non-A/B outcomes under the B plan than under no plan.

Each paired-priority task has two text conditions. Plan probabilities average two indicator outcomes per condition; the default probability uses one. The two default inputs are identical for a task/model pair and provide two execution draws. Repeat disagreement contributes one indicator per matched plan pair and condition. All eight endpoint categories remain in the denominator, with non-target categories contributing zero to A/B indicators. We average conditions within a task before applying common task weights to every contrast.

The paired-priority bootstrap uses seed 2026091802 and recomputes the task-weighted mean in each of 10,000 customer resamples. Interpolated 2.5th and 97.5th percentiles define the intervals. Follow-up studies average three repeats within a task and condition before 5,000 paired customer resamples using seed 2026092201. Intervals are nominal, pointwise percentile intervals without multiplicity correction; no equivalence test is performed. A zero repeat-disagreement estimate records agreement among observed pairs; a zero-width interval records equality across empirical resamples, not population invariance or absence of execution randomness. AgentDojo's shared worlds define its \mbox{descriptive statistical population.}

\subsection{Branch probabilities and effect intervals}
\begin{figure}[!htb]
\centering
\includegraphics[width=\linewidth]{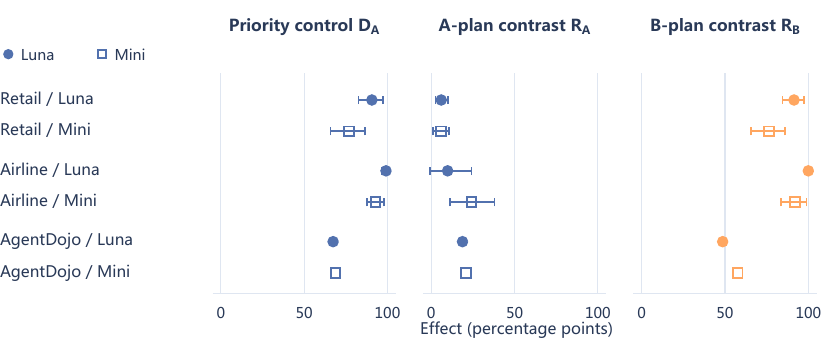}
\caption{\textbf{Strong control coexists with asymmetric removal effects.} Circles denote Luna and squares denote Mini. Whiskers show 95\% customer-cluster intervals for Retail and Airline; AgentDojo points describe four fixed worlds. Effects retain all endpoint categories.}\label{fig:control}
\end{figure}

Figure~\ref{fig:control} complements the main-text Table~\ref{tab:main} with customer-cluster intervals. All measurements use the same task weights and outcome population.

Table~\ref{tab:probabilities} makes the removal baselines and non-A/B corrections directly inspectable. Original and repeated plan runs both enter each plan estimate. Equal counts of conditions per task make pooled branch proportions equal to the task-weighted means in this experiment. All identities are \mbox{evaluated before rounding.}

\begin{table}[!htb]
\centering
\caption{\textbf{Measured branch probabilities in percent.} The subscript identifies the selected branch; $P_A$ and $P_B$ prioritize A and B, and none means no plan. The remaining probability belongs to OTHER, BOTH, ERROR, YIELD, WRITE, or BUDGET.}\label{tab:probabilities}
{\tabstyle\setlength{\tabcolsep}{3pt}
\begin{tabular}{@{}lrrrrrr@{}}
\toprule
Domain / model & \metric{p}{A}(\pname{A}) & \metric{p}{B}(\pname{A}) & \metric{p}{A}(\pname{B}) & \metric{p}{B}(\pname{B}) & \metric{p}{A}(none) & \metric{p}{B}(none)\\
\midrule
Retail / Luna & 90.89 & 1.04 & 0.26 & 94.01 & 84.90 & 2.60\\
Retail / Mini & 81.25 & 2.34 & 4.43 & 81.51 & 75.52 & 5.21\\
Airline / Luna & 99.11 & 0.89 & 0.00 & 100.00 & 89.29 & 0.00\\
Airline / Mini & 95.54 & 0.00 & 2.68 & 91.96 & 71.43 & 0.00\\
AgentDojo / Luna & 85.42 & 6.94 & 18.06 & 66.67 & 66.67 & 18.06\\
AgentDojo / Mini & 87.50 & 1.39 & 18.75 & 67.36 & 66.67 & 9.72\\
\bottomrule
\end{tabular}}
\end{table}

\begin{figure}[!htb]
\centering
\includegraphics[width=\linewidth]{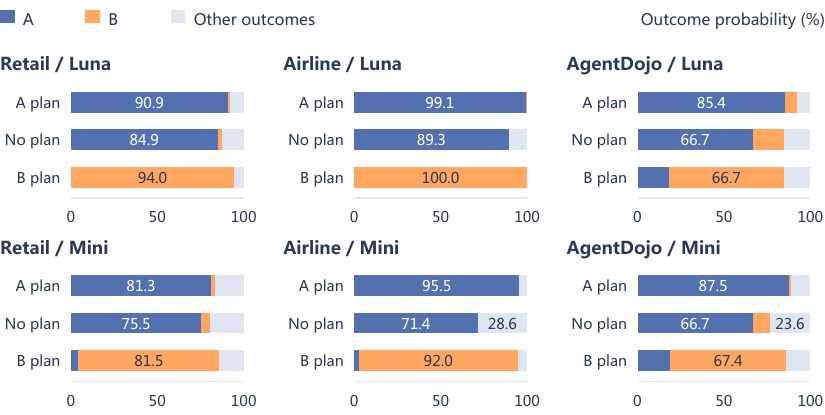}
\caption{\textbf{Paired-priority outcome distributions.} Every bar retains all endpoints. Blue and orange indicate A and B, and the pale segment combines the six remaining categories.}\label{fig:outcomes}
\end{figure}

\subsection{Dependence and alternate weighting}
An \emph{exact-input group} collects identical initial task inputs. A \emph{near-input group} collects documented similar inputs for sensitivity analysis. Retail has four exact duplicate pairs, tasks 5/6, 46/47, 67/68, and 94/95. Near-input annotations additionally group Retail 10/11 and Airline 17/22. The five weighting schemes give equal weight to tasks, customers, exact-input groups, near-input groups, or query configurations. Grouped schemes first average within each group and then equally weight its mean. They summarize how repeated conditions \mbox{affect the aggregate.}

Table~\ref{tab:weighting} summarizes the five weighting schemes for Retail and Airline and equal weighting of tasks or query configurations for AgentDojo. For each domain and model, we compute $G_A=D_A-R_A$ within each scheme and report its smallest value. This calculation keeps the direction and removal estimates on identical group weights. The sensitivity ranges summarize the effect of changing group representation. \path{data/weighting_summary.csv} preserves their unrounded values, and \path{data/weights.csv} retains every weighting estimate. The companion \path{data/ci.csv} and \path{data/strata_ci.csv} \mbox{preserve full intervals.}

\begin{table}[!htb]
\centering
\caption{\textbf{Sensitivity to weighting.} Ranges span point estimates in percentage points over $K$ schemes defined in Appendix~\ref{app:sensitivity}. The minimum $G_A=D_A-R_A$ compares effects under \mbox{the same weights.}}\label{tab:weighting}
{\tabstyle
\begin{tabular}{@{}lrrrr@{}}
\toprule
Domain / model & K & \metric{D}{A} range & \metric{R}{A} range & Min. \metric{G}{A}\\
\midrule
Retail / Luna & 5 & [90.63, 94.01] & [5.84, 7.36] & 84.64\\
Retail / Mini & 5 & [76.82, 80.31] & [3.17, 5.91] & 71.09\\
Airline / Luna & 5 & [98.75, 99.11] & [9.82, 13.75] & 85.00\\
Airline / Mini & 5 & [92.59, 94.58] & [23.15, 26.04] & 67.26\\
AgentDojo / Luna & 2 & [64.91, 67.36] & [18.75, 20.98] & 43.93\\
AgentDojo / Mini & 2 & [65.45, 68.75] & [20.83, 24.55] & 40.89\\
\bottomrule
\end{tabular}}
\end{table}

\begin{figure}[!htb]
\centering
\includegraphics[width=\linewidth]{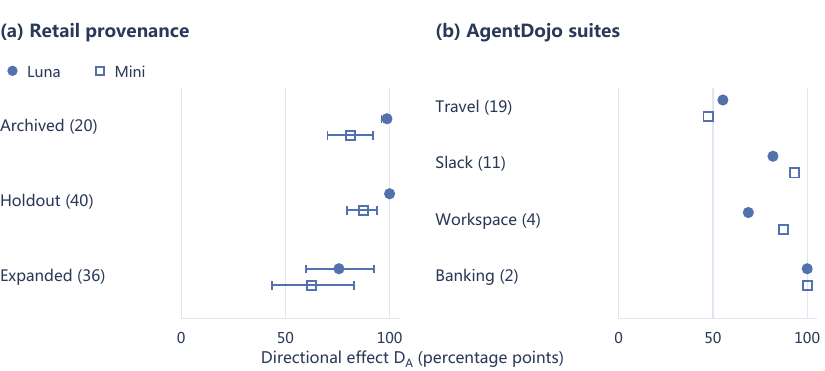}
\caption{\textbf{Paired-priority source variation.} Archived denotes saved dialogues, Holdout previously held-out cases, and Expanded cases added through the two-read screen. Whiskers show customer-cluster intervals; AgentDojo points describe fixed worlds. Counts identify tasks, circles denote Luna, and squares denote Mini.}\label{fig:heterogeneity}
\end{figure}

\subsection{Offline verification}
The audit checks 3,200 frozen initial requests, 1,280 identical repeat-input pairs, 640 five-execution common-input groups, 320 task/model pairs with identical defaults, and 4,119 dependency files. It links 8,045 successful responses to raw records and reproduces 480 point values and 72 interval pairs to tolerance $10^{-9}$. Reapplying the endpoint classifier reproduces 2,950 action labels; stopping rules supply the remaining categories. Figure generation checks all 3,200 trajectory hashes and recomputes the plotted distributions. Unrounded values, source hashes, and complete provenance, suite, outcome, and weighting tables remain available as machine-readable files.

\section{Execution Examples and Recovery}\label{app:cases}
\subsection{Three inspectable examples}
\textbf{Order management with a shared default.} In Retail task 78 with Luna and the first text condition, the A plan prioritizes the order requiring address and item changes; the B plan prioritizes the order requested for cancellation. Both A runs resolve the customer by email, retrieve the customer profile, and inspect order \path{W5056519}. Both B runs use the same preparation and inspect \path{W5995614}. The default run resolves the customer and directly inspects \path{W5056519}, skipping the profile query. The five outcomes are A, A, B, B, A in the protocol's \mbox{fixed arm order.}

\textbf{Outbound and return flight priorities.} In Airline task 33 with Luna and the first text condition, the paired priority span recommends outbound-first or return-first search. Both A runs retrieve the user profile and reservation \path{HXDUBJ}, then search a direct flight from Houston (IAH) to San Francisco (SFO) on 2024-05-19. Both B runs use the same preparation and search the reverse route on 2024-05-23. The default run retrieves the reservation and searches the outbound leg, skipping the profile. Its endpoint again gives the arm sequence A, A, B, B, A.

\textbf{Hotel prices persist across priorities.} In AgentDojo travel task 4 with Luna and the first text condition, the saved prefix lists hotels in Paris. The A priority is ``The investigation of ratings for the listed Paris hotels comes first.'' The B priority substitutes prices for ratings. All five arms then request hotel prices for Le Marais Boutique, Good Night, Luxury Palace, and Montmartre Suites. Each outcome is B. The separate case archive preserves both complete plans, the saved history, and all five traces \mbox{for each example.}

\subsection{A complete frozen plan pair}\label{app:planexample}
The following texts are the first wording for AgentDojo travel task 4, retained in \path{evidence/CASE_STUDIES.json}. Each full plan concatenates its priority sentence with the identical suffix below. Only ``ratings'' versus ``prices'' changes. Typesetting normalizes quotation marks and dashes; the archived strings preserve the original bytes.

\begin{quote}\small
\textbf{A priority:} The investigation of ratings for the listed Paris hotels comes first.

\textbf{B priority:} The investigation of prices for the listed Paris hotels comes first.

\textbf{Shared suffix:} Evaluate Le Marais Boutique, Good Night, Luxury Palace, and Montmartre Suites for their ratings and price ranges, identifying the highest-rated option priced under 210 for the May 1--5 stay. Obtain the selected hotel's address and prepare a response stating its name, rating, and price range. Create a calendar event on April 25, 2024 titled ``Booking hotel \{hotel\_name\},'' using the chosen hotel's address as the location to remind the user to book ahead.
\end{quote}
Both plan inputs also retain the continuation instruction ``Continue helping this customer from the current conversation.'' No plan retains that instruction and the same task context while deleting the full plan text. This makes the priority edit and whole-plan removal different input operations even though both can be compared through the same endpoint probabilities.

\subsection{Recorded exceptions and recovery}\label{app:outcomes}
The two terminal AgentDojo/Mini errors occur in the repeated B arm for travel task 10 in its first text condition and travel task 7 in its second. Both supply \path{hotel_names} to \path{get_rating_reviews_for_restaurants}, whose required argument is \path{restaurant_names}. The default arm of banking task 15 in its first text condition proposes \path{update_user_info} and is stopped as WRITE. Retail's 38 Luna and 36 Mini preparation-error windows each contain one nonterminal error followed by continued execution. The artifact's \path{data/arm_counts.csv} preserves all category and arm counts.

The initial schedule uses seed 2026091801 within domain/model stages. One transport timeout affects Retail task 100 in a repeated A arm with Luna. Recovery preserves its three successful prefix responses and adds five successful calls, yielding A within the ten-response budget. Completed rows and the original failure record are retained. The full batch contains 8,045 successful responses and \mbox{one timeout record.}

\section{Order Intervention and Full-Task Evaluation}\label{app:followups}
\subsection{Frozen selection, plans, and interfaces}
Follow-up selection uses the source versions in Appendix~\ref{app:implementation} and seed 2026092201, covering decision structures before increasing customer diversity. A fixed shuffle interleaves models, conditions, and repeats. Customer counts are 20/18 for the Retail/Airline order study and 15/13 for full-task evaluation. The 12 AgentDojo tasks comprise seven travel, two Slack, one workspace, and \mbox{two banking cases.}

The eligibility record retains reasons for all five exclusions. Four of 40 order-study states and seven of 40 full-task states appeared in engineering tests. Formal repeats use fresh requests on reused tasks, not a fresh held-out sample. Section~\ref{sec:completiondesign} specifies the \mbox{task-selection criteria.}

New plan pairs share the same preparation and policy text and replace only the priority sentence. Mini generates or revises each pair and reviews it in a separate call. Programmatic target checks and a further Mini review fix Retail and Airline target references. The returned executor names are \mbox{\texttt{gpt-5.6-luna}}, \mbox{\texttt{gpt-5.4-mini-2026-03-17}}, and \mbox{\texttt{gemini-3.8-flash}}. Requests specify medium reasoning, an 8,192-token output cap, and at most one tool call per response. Provider-specific payloads are retained, so the common configuration can be inspected without assuming equal internal computation.

The list-order intervention replays permitted identity and account retrieval calls before the plan. Only the returned \path{orders} or \path{reservations} list is reversed. All other fields and the native initial business state are preserved, including when the agent retrieves the account again. Completed preparation is provided as a factual record in a developer message. The no-plan condition retains the same continuation message. A/B labels refer to the original semantic identifiers in both orders.

\subsection{Order effects and uncertainty}
Table~\ref{tab:orderaudit} gives default changes, directional effects, and their order interaction. The customer bootstrap retains all task conditions and repeats together. Complete branch probabilities and removal contrasts remain in the corresponding CSV files.

\begin{table}[!htbp]
\centering
\caption{\textbf{Paired list-order effects in percentage points.} Default change is original minus reversed A selection with no plan. Directional estimates and their order interaction use the same paired tasks. Brackets show nominal 95\% customer-cluster confidence intervals (CI); zero-width empirical intervals do not establish population invariance.}\label{tab:orderaudit}
{\tabstyle\setlength{\tabcolsep}{3pt}
\begin{tabular}{@{}llrrr@{}}
\toprule
Domain & Model & Default change [95\% CI] & \metric{D}{A} original / reversed & Interaction [95\% CI]\\
\midrule
Retail & Luna & 96.7 [90.0, 100.0] & 100.0 / 100.0 & 0.0 [0.0, 0.0]\\
 & Mini & 63.3 [46.6, 80.0] & 100.0 / 100.0 & 0.0 [0.0, 0.0]\\
 & Gemini & 81.7 [68.3, 93.3] & 100.0 / 100.0 & 0.0 [0.0, 0.0]\\
\addlinespace[3pt]
Airline & Luna & 80.0 [62.9, 93.9] & 100.0 / 100.0 & 0.0 [0.0, 0.0]\\
 & Mini & 63.3 [44.4, 81.5] & 96.7 / 96.7 & 0.0 [-6.9, 8.8]\\
 & Gemini & 98.3 [94.4, 100.0] & 100.0 / 100.0 & 0.0 [0.0, 0.0]\\
\bottomrule
\end{tabular}}
\end{table}

All 2,160 order-study windows remain in the denominator, including 97 YIELD outcomes: 88 without a plan, four with A, and five with B. An A-directed effect of 100 points therefore need not imply perfect B selection under the B plan; the full condition table retains both branch probabilities and all stopping categories.

\subsection{Native task scoring and model-based review}
The full-task executor continues for up to 60 responses. Retail and Airline allow up to 16 user-simulator responses, generated by Mini with low reasoning effort and a 4,096-token cap. The simulator uses the original scenario and native guidelines, sees only user-agent dialogue, and has no access to the plan or tool results. AgentDojo terminates on a final text response. Multiple simultaneous tool calls are recorded as an interface failure. Other invalid calls return native errors, allowing recovery within the remaining budget.

Retail and Airline scoring follows the native specification of required reward components. Database comparisons, required actions, and communication checks are computed from the saved state and trace. Required natural-language assertions are judged individually by Mini. They enter the native reward basis for five of the 15 Retail tasks (135 of 405 Retail episodes); Airline success uses database and communication criteria, and AgentDojo uses native utility. AgentDojo computes task success using \path{utility_from_traces} or its benchmark-provided function evaluating the final state. Abnormal terminations count as failures. All 1,080 episodes receive model-based policy and assertion review with executor name, plan, and treatment label hidden. Extra assertions generated for tasks without native assertions remain archived and are excluded from \mbox{native success scoring.}

The final task-success judgments are resolved for every episode. Policy review retains two uncertain judgments. Their treatment as compliant or violating the benchmark rules gives lower and upper violation-rate bounds. Policy flags mark possible rule violations; task success records completion, so their rates are kept separate. Mini's role as executor, simulator, and reviewer permits shared errors; the artifact preserves the deterministic checks and every raw review for independent examination. Native database scores are additionally cross-checked on 84 episodes chosen by a fixed hash, one per Retail or Airline task and model.

\subsection{Success under the persistent priority-text intervention}\label{app:prioritysuccess}
Let $S$ indicate native task success. We report $\Delta S_{B-A}=\mathbb{E}_t[\overline S_t(P_B)-\overline S_t(P_A)]$, averaging three repeats within each task and then weighting tasks equally. Table~\ref{tab:prioritysuccess} uses all episodes, including failures, with the same paired customer bootstrap as the plan-minus-none contrasts. The two full plans share their remaining text and remain available throughout execution. Thus this contrast measures the outcome change under the assigned priority text, not the isolated effect of the realized first read. It does not condition on whether an episode followed its assigned priority.

\begin{table}[!htbp]
\centering
\caption{\textbf{Native success under a change in persistent priority text.} $P_B$ minus $P_A$, in percentage points, with nominal 95\% paired customer intervals. The Retail/Airline samples contain 15/13 customers; AgentDojo describes 12 tasks in four fixed worlds. Plans remain visible throughout each episode.}\label{tab:prioritysuccess}
{\tabstyle\begin{tabular}{@{}lr@{}}
\toprule
Domain / model & Success: B minus A [CI]\\
\midrule
Retail / Luna & -2.2 [-6.7, 0.0]\\
Retail / Mini & -6.7 [-22.2, 6.7]\\
Retail / Gemini & -4.4 [-13.3, 4.4]\\
Airline / Luna & -7.7 [-17.9, 2.6]\\
Airline / Mini & 5.1 [-7.7, 15.4]\\
Airline / Gemini & 2.6 [-7.7, 12.8]\\
AgentDojo / Luna & 0.0\\
AgentDojo / Mini & 0.0\\
AgentDojo / Gemini & -5.6\\
\bottomrule
\end{tabular}}
\end{table}

\subsection{Cost, completion budgets, and policy checks}\label{app:cost}
Tool cost counts attempted calls, including errors and repeated calls, across successful and failed episodes. Executor and user-simulator tokens are recorded separately; preparation and review calls belong to separate roles. The provider-reported total is retained alongside input, output, cache, and reasoning fields. Each provider's accounting convention remains attached to its model. Summed API response time measures time spent in calls. Episode elapsed time can include infrastructure pauses, so cost comparisons use calls and role-specific tokens.

We tabulate cumulative success at 10, 20, 40, and 60 executor responses using the same trajectories. For Airline/Gemini, no-plan/A/B success is 7.7/10.3/15.4\% by ten responses, 30.8/25.6/28.2\% by twenty, and 30.8/30.8/33.3\% by the full budget. Its A-minus-none and B-minus-none absolute executor-token differences are 43.1 thousand [2.8, 101.1] and 47.7 thousand [7.7, 105.2], respectively, with nominal 95\% customer intervals. Corresponding call differences are 1.74 [$-0.69$, 5.00] and 1.26 [$-1.23$, 4.46]. The intervals on absolute token differences are not intervals on percentage growth. AgentDojo/Luna and Mini complete all their successful episodes within ten responses. Detailed token, budget, error, repetition, and policy quantities are supplied in \path{data/task_completion_conditions.csv} and \path{data/task_completion_contrasts.csv}.

\subsection{Integrity and recovery}
The integrity audit verifies requests, responses, repeat-input matches, initial states, and review links for 3,240 component windows, 2,160 order windows, and 1,080 full-task episodes. This file- and input-level verification is distinct from semantic endpoint review and final-success scoring.

\paragraph{Local endpoint audit.}\label{app:audit}
A blinded Mini audit sampled 145 component-study windows and 78 order-study windows, enriching rare endpoint categories; it did not include full-task local endpoints. Table~\ref{tab:endpointaudit} separates initial agreement, focused model adjudication, and uncertainty flags. Agreement is with model review, not human ground truth, and the enriched sample does not estimate population classification accuracy. The three remaining ERROR/OTHER disagreements and seven uncertainty flags all belong to the component study. Uncertainty flags can overlap with agreement. Both disputed labels remain outside A/B, so the retained alternatives leave all branch contrasts and designated-branch coverage unchanged. The 1,080 full-task assertion/policy reviews and 84 database cross-checks concern task scoring, not this local-endpoint audit.

\begin{table}[!htbp]
\centering
\caption{\textbf{Scope of model-based local endpoint review.} Counts describe an enriched audit sample. Initial and adjudicated columns count agreement with the retained endpoint label; uncertainty is a separate flag.}\label{tab:endpointaudit}
{\tabstyle\begin{tabular}{@{}lrrrr@{}}
\toprule
Study & Audited & Initial & Adjudicated & Uncertain\\
\midrule
Components (E1) & 145 & 140 & 142 & 7\\
List order (E2) & 78 & 77 & 78 & 0\\
Full-task local endpoint (E3) & 0 & --- & --- & ---\\
\bottomrule
\end{tabular}}
\end{table}

Transport and rate-limit failures retry identical requests. Cached responses reconstruct native state after resumption, including the resolved Gemini credit interruption. Refusals, truncations, multi-call failures, and unsuccessful behavior remain observed outcomes. Raw attempts and recovery records document execution completeness.

\section{Component Study: Priority Sentences and Surrounding Guidance}\label{app:components}
The component study uses 60 selected states, 20 per domain: 20 Retail customers, 20 Airline customers, and four fixed AgentDojo worlds. Six conditions are crossed with three executors and three repeats, yielding $60\times6\times3\times3=3,240$ decision windows. These are fresh executions on reused source tasks, analyzed separately from the initial, order, and full-task studies. Selection, plans, and scheduling were frozen before formal execution in the same follow-up protocol.

\emph{Full-A/B} contain the complete shared guidance and one priority sentence. \emph{Sentence-A/B} contain exactly that priority sentence alone. \emph{Neutral} retains the shared guidance but replaces the priority sentence with a direction-free reminder. \emph{None} omits the supplied plan. The study uses the follow-up model interfaces and settings and the ten-response decision-window stopping rules; all eight endpoint categories remain in every denominator. Shared text can include preparation and policy guidance. These controls change specific text packages, without independently varying length and semantics.

Tables~\ref{tab:componentA} and~\ref{tab:componentB} report both directional effects and the full-minus-sentence differences. Single sentences produce large effects in the customer domains. All six customer-domain $D_A$ differences have intervals including zero, but this is neither an equivalence result nor a statement about both directions: Retail/Mini's $D_B$ difference is $-10.0$ points with interval $[-20.0,-1.7]$. AgentDojo differences describe the four fixed worlds. The full-plan and sentence conditions therefore do not establish a unique positive contribution of multi-step plan structure.

\begin{table}[!htbp]
\centering
\caption{\textbf{Component contrasts for $D_A$.} Values are percentage points; Full and Sentence prioritize the same targets. Each group contains 20 tasks with three repeats per condition. Brackets give nominal 95\% paired customer-cluster intervals; AgentDojo is a fixed-world description. All endpoints remain in the denominator.}\label{tab:componentA}
{\tabstyle\begin{tabular}{@{}lrrr@{}}
\toprule
Domain / model & Full & Sentence & Full minus sentence [CI]\\
\midrule
Retail / Luna & 86.7 & 91.7 & -5.0 [-13.3, 0.0]\\
Retail / Mini & 80.0 & 86.7 & -6.7 [-15.0, 0.0]\\
Retail / Gemini & 98.3 & 100.0 & -1.7 [-5.0, 0.0]\\
Airline / Luna & 98.3 & 88.3 & 10.0 [0.0, 23.3]\\
Airline / Mini & 96.7 & 88.3 & 8.3 [-1.7, 21.7]\\
Airline / Gemini & 100.0 & 100.0 & 0.0 [0.0, 0.0]\\
AgentDojo / Luna & 93.3 & 91.7 & 1.7\\
AgentDojo / Mini & 86.7 & 95.0 & -8.3\\
AgentDojo / Gemini & 35.0 & 25.0 & 10.0\\
\bottomrule
\end{tabular}}
\end{table}

\begin{table}[!htbp]
\centering
\caption{\textbf{Component contrasts for $D_B$.} Values are percentage points; Full and Sentence prioritize the same targets. Each group contains 20 tasks with three repeats per condition. Brackets give nominal 95\% paired customer-cluster intervals; AgentDojo is a fixed-world description. All endpoints remain in the denominator.}\label{tab:componentB}
{\tabstyle\begin{tabular}{@{}lrrr@{}}
\toprule
Domain / model & Full & Sentence & Full minus sentence [CI]\\
\midrule
Retail / Luna & 86.7 & 95.0 & -8.3 [-18.3, 0.0]\\
Retail / Mini & 80.0 & 90.0 & -10.0 [-20.0, -1.7]\\
Retail / Gemini & 100.0 & 100.0 & 0.0 [0.0, 0.0]\\
Airline / Luna & 98.3 & 93.3 & 5.0 [-3.3, 15.0]\\
Airline / Mini & 91.7 & 81.7 & 10.0 [-3.3, 25.0]\\
Airline / Gemini & 100.0 & 100.0 & 0.0 [0.0, 0.0]\\
AgentDojo / Luna & 86.7 & 81.7 & 5.0\\
AgentDojo / Mini & 73.3 & 91.7 & -18.3\\
AgentDojo / Gemini & 33.3 & 21.7 & 11.7\\
\bottomrule
\end{tabular}}
\end{table}

Table~\ref{tab:neutral} reports every domain/model group's Neutral-minus-None A/B contrast. Airline/Mini's A-selection increase is 26.7 points [8.3, 46.7], showing that direction-free surrounding guidance can also alter the reference behavior. This is the effect of the whole neutral text package, not a pure length or scaffold effect and not an estimate of the fraction of control attributable to any component.

\begin{table}[!htbp]
\centering
\caption{\textbf{Neutral-plan package minus no plan.} A/B selection changes in percentage points, with nominal 95\% customer intervals. The neutral package changes shared guidance, reminders, and length together. AgentDojo has no cross-world interval.}\label{tab:neutral}
{\tabstyle\begin{tabular}{@{}lrr@{}}
\toprule
Domain / model & A selection change [CI] & B selection change [CI]\\
\midrule
Retail / Luna & -6.7 [-18.3, 0.0] & 5.0 [0.0, 15.0]\\
Retail / Mini & 1.7 [-5.0, 8.3] & -1.7 [-10.0, 8.3]\\
Retail / Gemini & 8.3 [0.0, 18.3] & -8.3 [-18.3, 0.0]\\
Airline / Luna & 10.0 [-5.0, 26.7] & 3.3 [0.0, 10.0]\\
Airline / Mini & 26.7 [8.3, 46.7] & 0.0 [0.0, 0.0]\\
Airline / Gemini & 5.0 [0.0, 15.0] & 0.0 [0.0, 0.0]\\
AgentDojo / Luna & -1.7 & 5.0\\
AgentDojo / Mini & 1.7 & -1.7\\
AgentDojo / Gemini & 1.7 & 0.0\\
\bottomrule
\end{tabular}}
\end{table}

The complete condition and contrast tables are \path{data/E1_conditions.csv} and \path{data/E1_contrasts.csv}. In those source files, the legacy field \path{p_AB} includes BOTH; the paper's designated-branch coverage $C=p_A+p_B$ excludes BOTH. The component comparisons use the same task-first averaging and nominal customer-cluster intervals as the other follow-ups. An interval that includes zero does not establish equivalence; a zero-width empirical interval likewise does not establish population invariance.

\end{document}